\documentclass[10pt]{article}

\usepackage[preprint]{tmlr}
\usepackage{amsmath,amssymb}
\usepackage{booktabs}
\usepackage{array}
\usepackage{enumitem}
\usepackage{graphicx}
\usepackage{xcolor}
\usepackage{multirow}
\usepackage{caption}
\usepackage{subcaption}
\usepackage{hyperref}
\usepackage{microtype}

\title{Emergent Sparsity in Frozen Random CNN Feature\\ Extractors for Deep Reinforcement Learning}

\author{\name Scott M. Norton \email nortonsm@urgrad.rochester.edu \\
        \addr Independent Researcher}

\def\month{04}
\def\year{2026}
\def\openreview{\url{https://openreview.net/forum?id=XXXX}}

\begin{document}

\maketitle

\begin{abstract}
We report a striking phenomenon: deep reinforcement learning agents trained with \textbf{frozen, randomly initialized} CNN feature extractors spontaneously develop extremely sparse fully-connected representations, without any sparsity-inducing objective. In the first fully-connected layer (FC1, $3{,}136 \to 64$), agents compress all task-relevant information through as few as 1--3 neurons out of 64 for deterministic Pong (and 5--11 for Pong trained under sticky-action stochasticity), while trainable CNNs activate 55--64 neurons under matched conditions. We establish four principal findings.

First, \textbf{FC1 sparsity scales with task complexity across games}: 1--11 for Pong, 19--26 for Breakout, $\sim$42 for Space Invaders, with width-scaling experiments confirming this reflects task structure rather than a fixed fraction of network capacity. Second, \textbf{within-game scaling at matched hyperparameters} sharpens this: three same-recipe frozen Pong seeds produce 5, 7, and 11 active neurons; the 5-neuron seed plateaus at $+14$ reward while the 7- and 11-neuron seeds reach expert ($+18.4$, $+18.7$). With $n=3$ this should be read as suggestive rather than establishing a smooth scaling law, but it is consistent with the random projection's usable dimensionality setting a ceiling on achievable performance at fixed task and capacity. Third, \textbf{ablation confirms necessity}: removing the active neurons crashes performance to random play across two PPO implementations and four games. Fourth, \textbf{the information bottleneck is committed early}: a 47-checkpoint sweep across three Pong seeds shows the active set locked by 15--30M steps with the ablation signature already present at 5M, while reward turns positive 35--105M steps later, with the lag widening as bottleneck capacity tightens ($n=3$, ordinal pattern). The active-set commits-early property recurs in trainable agents, though with a qualitatively different content (which large fraction participates rather than which small set). A complementary cross-regime finding: in Breakout, frozen and trainable CNNs reach competitive reward via structurally different bottleneck signatures. Frozen agents use 17--25 active neurons at participation ratio $\sim$10--14 (FC2-output activations); trainable agents use 51 at participation ratio $\sim$3.6. Active count and participation ratio are complementary indicators of representational compression. Beyond RL: wherever input dimensionality dwarfs intrinsic task dimensionality, gradient descent on a frozen random projection may reveal the effective rank of the underlying problem without explicit sparsity machinery.
\end{abstract}

\section{Introduction}
\label{sec:intro}

Deep reinforcement learning (DRL) operates under a fundamental tension between capacity and structure. Modern agents deploy neural networks with millions of parameters to solve environments governed by a handful of state variables. The game of \textit{Pong} (a paddle game in which two players return a ball across a court) is fully described by the positions and velocities of the ball and two paddles, roughly five continuous variables. Yet standard practice involves training end-to-end convolutional neural networks (CNNs) that map high-dimensional pixel inputs to actions through dense, distributed computation. While overparameterized models are effective, they obscure the underlying simplicity of the task. They learn to solve the problem, but they do not reveal the \textit{structure} of the problem.

Recent work has increasingly documented pathologies arising from this overparameterization: representation rank collapse degrades effective feature dimensionality \citep{kumar2021implicit}, dormant neurons accumulate and reduce capacity \citep{sokar2023dormant}, plasticity loss impairs the ability to learn new behaviors \citep{lyle2023understanding, dohare2024loss}, and the primacy bias causes over-commitment to early representations \citep{nikishin2022primacy}. These phenomena share a common thread: end-to-end training creates optimization dynamics that interact with representation learning in complex, often counterproductive ways.

The prevailing approach to sparsity treats it as something that must be \textit{imposed}. Pruning removes parameters after training \citep{han2015learning, lecun1990optimal}. $\ell_1$ regularization penalizes dense solutions \citep{tibshirani1996regression}. The lottery ticket hypothesis reveals sparse subnetworks within dense initializations, but finding them requires iterative magnitude pruning \citep{frankle2019lottery}. Even the strong lottery ticket hypothesis, which proves that random networks contain performant subnetworks without weight training, requires explicit search via masking algorithms \citep{ramanujan2020hidden, malach2020proving}. In all cases, sparsity is engineered, not emergent.

In this work, we demonstrate that under a specific architectural constraint---a \textbf{frozen, randomly initialized CNN feature extractor}---sparsity is not engineered but \textit{emerges} from learning dynamics. When a PPO agent must construct a policy from a fixed basis of random features, it spontaneously converges to solutions utilizing a small fraction of available neurons. In the extreme case of deterministic Pong, the agent ignores over 95\% of available capacity, solving the task with as few as 1--3 active neurons out of 64. Throughout this work we use sticky actions \citep{machado2018revisiting} where appropriate---a standard Atari Learning Environment stochasticity protocol in which the agent's previous action is repeated with probability $p_\text{sticky} = 0.25$ at each time step---to ensure findings reflect genuine reactive control rather than memorization of fixed action sequences. We establish four main contributions.

\begin{enumerate}[leftmargin=*,itemsep=0.2em]
  \item \textbf{Cross-game sparsity scales with task complexity:} 1--11 active for Pong, 19--26 for Breakout, $\sim$42 for Space Invaders. Width-scaling experiments confirm this reflects task structure, not capacity (Section~\ref{sec:results}).

  \item \textbf{Within-game capacity-limited scaling at matched hyperparameters:} three same-recipe frozen Pong seeds produce 5/7/11 active neurons; the 5-neuron seed reaches $+14.1$ while the 7- and 11-neuron seeds reach expert ($+18.4$, $+18.7$). With $n=3$, this is a suggestive pattern rather than a fitted scaling law, but it is consistent with the random projection's usable dimensionality setting a ceiling on performance at fixed task and capacity (Section~\ref{sec:within-game}).

  \item \textbf{Ablation-confirmed bottleneck:} the active set is necessary (FC1-Remove crashes to random play across two PPO implementations and four games) and sufficient (Section~\ref{sec:ablation}).

  \item \textbf{Early commitment of the bottleneck precedes performance:} a 47-checkpoint sweep across three Pong seeds shows FC1-Remove $= -21$ from 5M and active count locked by 15--30M; reward turns positive 35--105M steps later, with the lag widening as bottleneck capacity tightens across the three seeds. Trainable agents also exhibit early active-set commitment, though to a qualitatively different content---a broad set rather than a narrow one---and with different post-commitment dynamics (Section~\ref{sec:trajectory}).
\end{enumerate}

A complementary cross-regime finding: in Breakout, frozen and trainable CNNs reach competitive reward via structurally different bottleneck signatures. Frozen agents use 17--25 active neurons at FC2-output participation ratio $\sim$10--14; trainable agents use 51 at participation ratio $\sim$3.6. Active count and participation ratio are complementary indicators of representational compression (Section~\ref{sec:dimensionality}).

\begin{figure}[ht]
  \centering
  \includegraphics[width=0.85\linewidth]{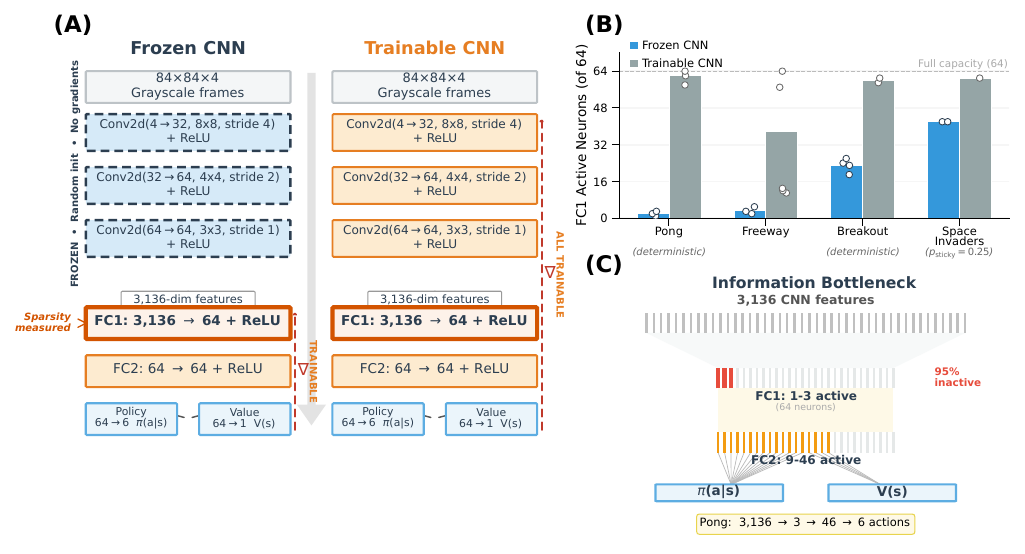}
  \caption{Frozen random CNN architecture and the cross-game sparsity it produces. \textbf{(A)}~Architecture schematic: in the frozen condition the CNN is initialized randomly and never updated; only the FC layers and policy/value heads receive gradients. \textbf{(B)}~Across four games at their training stochasticity regime---Pong (deterministic), Freeway (deterministic-aligned ablation-verified seeds), Breakout (deterministic), Space Invaders ($p_\text{sticky}=0.25$)---frozen-CNN agents (blue) recruit dramatically fewer FC1 neurons than trainable controls (gray), and the gap widens for simpler games. Pong frozen counts under stochastic sticky-action training are higher (5--11) and reported separately in Section~\ref{sec:within-game} and Figure~\ref{fig:ablation}. \textbf{(C)}~Schematic of the resulting two-stage information bottleneck: 3{,}136 CNN features feed a 1--3 active FC1 subset (Pong example), which feeds a 9--46 active FC2-output subset, which feeds the policy and value heads.}
  \label{fig:architecture}
\end{figure}

\section{Related Work}
\label{sec:related}

\paragraph{Sparsity in neural networks.} Across all established approaches, sparsity is \textit{explicitly induced}. Pruning removes parameters based on magnitude or gradient criteria \citep{lecun1990optimal, han2015learning, molchanov2017pruning}. The lottery ticket hypothesis \citep{frankle2019lottery} showed that pruned networks can match dense performance when reset to initial weights, and the strong lottery ticket hypothesis proved that random networks contain performant subnetworks without any weight training, but finding them requires explicit search via masking \citep{ramanujan2020hidden, malach2020proving, gadhikar2024partial}. In RL specifically, \citet{obandoceron2024pruned} achieved 90--95\% weight sparsity with \textit{improved} Atari performance through gradual magnitude pruning, and \citet{ma2025sparsity} showed that even random pruning before training outperforms dense DRL networks. \citet{davelouis2025interplay} studied designed sparse structures in fixed-weight RL networks. Combined, these results suggest massive overparameterization in DRL but all approach sparsity from the ``how much can we remove?'' direction. We ask the complementary question: ``how little does gradient descent \textit{choose to use} when features are fixed?'' Sparsity in our setting emerges spontaneously from standard PPO when the CNN is frozen---no pruning, no regularization, no architectural constraints.

\paragraph{Reservoir computing and random features.} Our frozen-CNN architecture (a fixed nonlinear projection followed by a trainable readout) is formally a reservoir computing system \citep{jaeger2001echo, maass2002real}. \citet{chang2020reinforcement} applied convolutional reservoir computing to Atari using CMA-ES, but did not analyze activation sparsity. The theoretical foundation for random feature models was established by \citet{rahimi2007random}; more recently, \citet{coil2025frozen} found that frozen random layers act as implicit regularizers, and \citet{zhang2025pretrained} found that frozen random features provide surprisingly competitive baselines for visual RL. Related work by \citet{gaier2019weight} showed that Weight Agnostic Neural Networks can solve RL tasks with random shared weights, and \citet{cuccu2019playing} achieved competitive Atari performance with as few as 6--18 neurons via neuroevolution. None of these works examine how activation sparsity in the readout tracks task complexity.

\paragraph{Intrinsic dimensionality and representation dynamics.} \citet{li2018measuring} measured the minimum parameter degrees of freedom needed to solve RL tasks, reporting $d_{\text{int}} \approx 6{,}000$--$10{,}000$ for Pong. Our finding that 3 active neurons correspond to ${\sim}9{,}400$ effective parameters ($3 \times 3{,}136$ input weights) is strikingly consistent. \citet{sokar2023dormant} discovered that neurons in DRL networks progressively become dormant during training and explicitly identified the relationship between task complexity and dormancy as \textbf{open future work}. Our findings provide a direct answer: under frozen random features, the number of active neurons tracks task complexity, while dormant neurons emerge as a byproduct of discovering minimal sufficient representations. The plasticity loss literature \citep{lyle2023understanding, dohare2024loss, abbas2023plasticity} documents how permanently inactive units increase during training. \citet{moalla2024collapse} identified a bidirectional feedback loop between representation collapse and trust-region degradation. Our work reframes these phenomena: in frozen-CNN agents, low FC-layer dimensionality is \textit{informative}, reflecting task structure rather than pathological collapse.

\section{Methods}
\label{sec:methods}

\subsection{Architecture}

We study the Nature-DQN CNN architecture \citep{mnih2015human}: three convolutional layers (32 filters $8 \times 8$ stride 4, 64 filters $4 \times 4$ stride 2, 64 filters $3 \times 3$ stride 1) producing 3,136-dimensional feature vectors, followed by two fully connected layers (FC1: $3{,}136 \to 64$, FC2: $64 \to 64$) with ReLU activations, and separate policy and value heads. In the \textbf{frozen-CNN condition}, CNN weights are initialized with Kaiming normal initialization \citep{he2015delving} using a fixed seed and immediately frozen (\texttt{requires\_grad=False}). Only the FC layers, policy head, and value head are trainable (72.5\% of parameters); the CNN acts as a fixed random projection from pixel space to a 3,136-dimensional feature space. In the \textbf{trainable-CNN condition} (control), all layers are trained end-to-end with identical hyperparameters.

\subsection{Training Protocol}

All agents use Proximal Policy Optimization \citep{schulman2017proximal}. To establish robustness, we employ two independent PPO implementations: \textbf{Stable-Baselines3 (SB3)} \citep{raffin2021stable}, running synchronous PPO on both Mac and Ubuntu hardware, and \textbf{Sample Factory (SF)} \citep{petrenko2020sample}, running asynchronous PPO on Ubuntu. Both use standard Atari preprocessing (84$\times$84 grayscale, 4-frame stacking), learning rate $2.5 \times 10^{-4}$ with linear decay, clip range 0.1, and entropy coefficient 0.01. Hardware specifications and full hyperparameters appear in Appendix~\ref{app:details}.

\textbf{Stochasticity per experiment.} Sticky-action regimes vary by game and framework. \textbf{Pong frozen} runs span all three conditions: five SB3 seeds at 200M steps (deterministic), an extended SB3 800M run ($p_\text{sticky} = 0.25$ throughout), and three SB3 progressive-sticky seeds (c10, c20, c120) under a curriculum where $p_\text{sticky}$ interpolates linearly from $0$ at reward $-20$ to $0.25$ at reward $0$. \textbf{Pong trainable} runs (SB3 t1, t2, t3, t14) use the same progressive curriculum; SF Pong V5 trainable runs use fixed $p_\text{sticky} = 0.25$. \textbf{Breakout frozen} SB3 runs span two training regimes: three deterministic-trained seeds (seed5, seed10, seed0) and three progressive-sticky-trained runs (seed-10 lineage, reward-paced $p_\text{sticky}: 0\to 0.25$ schedule reaching competent play; Section~\ref{sec:limitations}); fixed $p_\text{sticky}=0.25$ Breakout training did not learn. SF Breakout trainable b1--b3 use fixed $p_\text{sticky} = 0.25$. \textbf{Space Invaders} (SB3) uses fixed $p_\text{sticky} = 0.25$. \textbf{Freeway} includes both sticky=$0$ and sticky=$0.25$ conditions. The progressive curriculum is methodologically load-bearing for stochastic frozen Pong (Appendix~\ref{app:crossframework}). Appendix~\ref{app:details} tabulates all configurations.

\subsection{Environments}
\label{sec:environments}

We study four Atari 2600 games spanning a range of intrinsic complexity: \textbf{Pong} ($\sim$5 state variables), \textbf{Breakout} ($\sim$15--25), \textbf{Space Invaders} ($\sim$40+), and \textbf{Freeway} ($\sim$10--15). All use standard ALE preprocessing. Our state-variable estimates are grounded in domain analysis and broadly consistent with the AtariARI semantic RAM annotations of \citet{anand2019unsupervised} (6 labeled variables for Pong, 10 for Breakout, 14 for Space Invaders); our higher estimates for Breakout and Space Invaders reflect that spatial patterns (brick layouts, alien grids) span multiple RAM bytes per labeled variable. Substituting AtariARI's labeled counts for the x-axis values in Figure~\ref{fig:scaling} would compress the slope of the frozen-CNN line but preserve rank ordering and the qualitative trend; the conclusion that frozen active counts \textit{track} task complexity is robust to the substitution. Freeway is a boundary case: pressing UP every frame scores $\sim$65\% of optimal reward without visual processing (Section~\ref{sec:freeway}). Full game details appear in Appendix~\ref{app:details}.

\subsection{Activation Analysis and Ablation Protocol}

We characterize the learned bottleneck using a custom activation analysis pipeline applied across 50--100 evaluation episodes and 1,000 activation samples. A neuron is classified as ``active'' if its mean post-ReLU activation exceeds 0.01. We measure effective dimensionality via PCA (PCA90/PCA95: components explaining 90\%/95\% of variance) and the \textbf{participation ratio}, $\text{PR} = (\sum_i \lambda_i)^2 / \sum_i \lambda_i^2$ where $\lambda_i$ are activation-covariance eigenvalues. PR is informative because two populations with the same active count can have very different PRs: in the trainable regime FC1 saturates at 64 but PR varies from $\sim$11 (expert) to $\sim$31 (stuck), making PR the discriminative metric.

The signature empirical test is \textbf{ablation verification}: for each model we evaluate baseline performance plus five masking conditions---FC1-Keep (zero inactive FC1; keep active), FC1-Remove (zero active FC1), FC2-Keep, FC2-Remove, and Both-Keep. An FC1-Remove crash to random play confirms a genuine information bottleneck; insensitivity indicates vestigial or bias-driven sparsity. Headline ablations use $n_{\text{eval}} = 100$ episodes; PR estimates at $n = 30$ can deviate by up to $\pm 5$ units at high-reward play (calibration in Appendix~\ref{app:n_eval}), and within-table orderings are preserved across this range.

\section{Results}
\label{sec:results}

\subsection{Cross-Game Sparsity Scaling}

\begin{figure}[ht]
  \centering
  \includegraphics[width=0.77\linewidth]{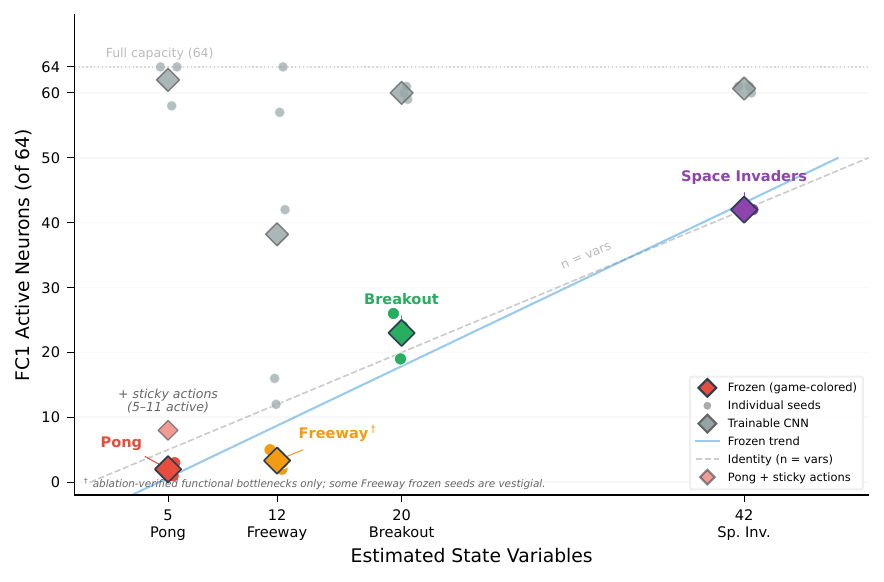}
  \caption{The frozen-CNN bottleneck size tracks task complexity, while trainable-CNN size does not. Active FC1 neuron counts (y-axis) vs.\ a \textit{heuristic estimate} of state variables per game (x-axis is ordinal, grounded in but not identical to AtariARI semantic RAM annotations; Section~\ref{sec:environments}); the ``Frozen trend'' line is a qualitative visual guide, not a fitted regression. Frozen-CNN agents (game-colored) approximately follow the identity line; trainable controls (gray) cluster near the 64-neuron capacity ceiling regardless of task. Frozen counts shown for the dominant training regime per game: Pong deterministic (1--3 active; sticky-action training shifts to 5--11, shown as the pink ``+ sticky actions'' diamond, Section~\ref{sec:within-game}), Freeway deterministic-aligned, Breakout deterministic, Space Invaders $p_\text{sticky}=0.25$. The Freeway frozen point ($\dagger$) shows only ablation-verified functional bottlenecks; some Freeway frozen seeds are vestigial (Appendix~\ref{app:freeway_complete}).}
  \label{fig:scaling}
\end{figure}

Our central result is summarized in Table~\ref{tab:main} and Figure~\ref{fig:scaling}. Across four Atari games of increasing complexity, frozen random CNNs produce dramatically sparser representations than trainable CNNs, with sparsity tracking intrinsic task complexity.

\begin{table}[ht]
\centering
\caption{Emergent FC1 sparsity across tasks (width = 64). Frozen-CNN active counts track estimated task complexity; trainable counts do not. Ranges reflect cross-seed variation. The Stochasticity column reports the training $p_\text{sticky}$ (det.\ $= 0$). Pong and Breakout frozen rows are split by training regime. $^\dagger$Freeway frozen reports only ablation-verified SB3 bottlenecks (Section~\ref{sec:freeway}); some Freeway seeds are vestigial. $^\ddagger$Breakout (sticky) frozen counts combine 3 deterministic-trained seeds (OOD eval at $p_\text{sticky}=0.25$) and 3 progressive-sticky-trained runs (in-distribution eval); see Table~\ref{tab:breakout_signatures}.}
\label{tab:main}
\small
\setlength{\tabcolsep}{4pt}
\begin{tabular}{lcccccc}
\toprule
Game & Stochasticity & Est.\ Vars & Frozen Active & Frozen \% & Train.\ Active & Train.\ \% \\
\midrule
\textbf{Pong} (det.)         & $p=0$          & $\sim$5      & 1--3   & 1.6--4.7\%  & 58--64 & 91--100\% \\
\textbf{Pong} (sticky-fixed) & $p=0.25$       & $\sim$5      & 3      & 4.7\%       & --     & --        \\
\textbf{Pong} (progressive)  & $p:0\to 0.25$  & $\sim$5      & 5--11  & 7.8--17.2\% & 63--64 & 98--100\% \\
\textbf{Breakout} (det.)     & $p=0$          & $\sim$20     & 19--26 & 30--41\%    & --     & --        \\
\textbf{Breakout} (sticky)$^\ddagger$ & $p=0.25$ & $\sim$20   & 17--25 & 27--39\%    & 52--59 & 81--92\%  \\
\textbf{Space Inv.}          & $p=0.25$       & $\sim$40+    & 42     & 65.6\%      & 61     & 95.3\%    \\
\textbf{Freeway}$^\dagger$   & $0$ / $0.25$   & $\sim$10--15 & 2--5   & 3--8\%      & 49--64 & 77--100\% \\
\bottomrule
\end{tabular}
\end{table}

The pattern is consistent across both frameworks (SB3 synchronous PPO and SF asynchronous PPO), across different hardware (Mac and Ubuntu/RTX~5090), and across training durations spanning 200M--853M environment frames. Trainable CNNs distribute information uniformly across available capacity; frozen CNNs, lacking the ability to co-adapt features with the readout, converge to solutions whose dimensionality reflects the task's intrinsic structure. Width-scaling experiments on Pong (widths 32, 64, 128) confirm the \textit{absolute} active count is stable (3--14 range regardless of width) while \textit{percentage} utilization varies inversely with width; seed variance exceeds width effects, ruling out a capacity-proportional explanation (Appendix~\ref{app:width}). Detailed per-game results appear in Appendices~\ref{app:pong_complete} and~\ref{app:freeway_complete}.

We highlight three frozen Pong runs that anchor the within-game analysis below: under the progressive sticky curriculum (Section~\ref{sec:methods}), three same-recipe SB3 seeds---here labelled c10, c20, c120\footnote{c10/c20/c120 are arbitrary CNN-init seed identifiers; the suffix is unrelated to step count or active-neuron count.}---reached stable expert/near-expert play with 11, 7, and 5 active neurons respectively. A separate 3-neuron SB3 frozen Pong solution (Seed~2, 200M, deterministic) supports a mechanistic decomposition into action-discriminating and value-encoding neurons (Appendix~\ref{app:mechanistic}, $R^2 = 0.993$ for value prediction).

\subsection{Ablation: The Active Set Is Necessary and Sufficient}
\label{sec:ablation}

The signature empirical test is whether the sparse active neurons carry \textit{all} task-relevant information. Table~\ref{tab:ablation} presents ablation results on Pong using the 6-seed dataset---three frozen seeds (c120, c20, c10) and three trainable seeds (t2, t14, t1)---all under the progressive sticky curriculum and $n_{\text{eval}} = 100$ evaluation. Because all six seeds use identical SB3 recipe and identical sticky regime, this table directly compares frozen and trainable bottleneck signatures.

\begin{table}[ht]
\centering
\caption{Ablation results across 6 matched-recipe SB3 Pong seeds (progressive sticky, $n_{\text{eval}} = 100$, all values from a single sweep dated 2026-05-02). FC1-Remove crashes all seeds to random play regardless of active count or frozen/trainable status. PR is the FC2-output participation ratio. Frozen seeds vary in active count and PR; trainable seeds saturate FC1 at 64 but vary in PR.}
\label{tab:ablation}
\small
\begin{tabular}{lcccccc}
\toprule
Run & FC1 act & FC2 out act & PCA95 & \textbf{PR} & Baseline & FC1-Rem \\
\midrule
\multicolumn{7}{l}{\textit{Frozen}} \\
c120 & 5  & 54 & 17 & \textbf{9.63}  & $+14.1 \pm 1.4$ & $-21.0$ \\
c20  & 7  & 46 & 23 & \textbf{11.74} & $+18.4 \pm 0.4$ & $-21.0$ \\
c10  & 11 & 57 & 26 & \textbf{13.73} & $+18.7 \pm 0.3$ & $-21.0$ \\
\midrule
\multicolumn{7}{l}{\textit{Trainable}} \\
t2  & 64 & 64 & 49 & \textbf{29.14} & $+14.4 \pm 1.2$ & $-21.0$ \\
t14 & 64 & 64 & 27 & \textbf{11.37} & $+20.6 \pm 0.1$ & $-21.0$ \\
t1  & 64 & 64 & 32 & \textbf{12.59} & $+20.5 \pm 0.1$ & $-21.0$ \\
\bottomrule
\end{tabular}
\end{table}

For Pong, FC1-Remove universally crashes performance to $-21.0$ regardless of active-neuron count (1, 2, 3, 5, 7, 11, or 64) or framework. The active set carries all task-relevant information; the inactive neurons are truly inert. Ablation results across additional Pong runs (SB3 800M; SF Kaiming-head and orthogonal-head variants) and Breakout (3 frozen seeds at deterministic evaluation, ablating to $\sim$2 in all three) appear in Appendices~\ref{app:pong_complete} and~\ref{app:freeway_complete}. As a cross-stochasticity robustness check, evaluating all three frozen Breakout headline seeds (seed5, seed10, seed0) under sticky-action evaluation ($p_\text{sticky} = 0.25$) preserves the active-neuron set ($|\Delta\text{active}| \le 4$ per seed) and produces consistent FC1-Remove crashes, even though baseline reward collapses by $41$--$93\%$ (the policy is calibrated to deterministic dynamics; the representation it reads from is not). Full $2\times 2$ data appear in Appendix~\ref{app:freeway_complete}. The universality of FC1-Remove $\to$ random play across these conditions establishes the bottleneck as a robust property of frozen-feature learning.

\begin{figure}[ht]
  \centering
  \includegraphics[width=0.85\linewidth]{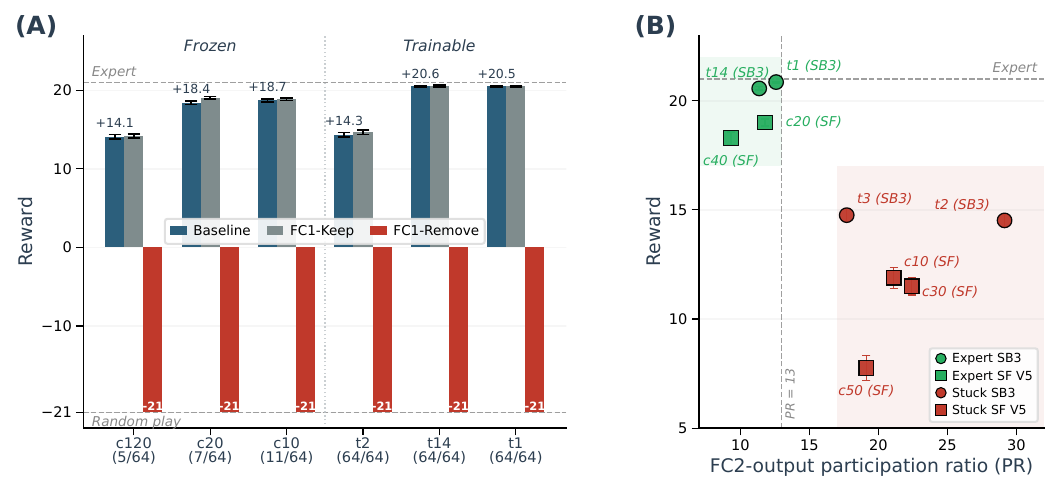}
  \caption{Ablation universality and a candidate cross-framework PR threshold. \textbf{(A)} Across 6 matched-recipe SB3 Pong seeds, Baseline (dark teal) and FC1-Keep (gray) bars are nearly identical---the active subset of FC1 neurons preserves all the policy's competence. The red FC1-Remove bars crash uniformly to random play, regardless of whether 5, 7, 11, or 64 neurons are active and regardless of frozen-vs-trainable status: \textit{the active subset is necessary across the entire bottleneck size range}. \textbf{(B)} Nine trainable Pong seeds spanning two PPO frameworks split into two clusters along the PR axis: 4 expert seeds (reward $\ge +18$, green) at PR$\le 12.6$ and 5 stuck seeds (reward $\le +15$, red) at PR$\ge 17.7$, separated by a 5.1-unit empty band. Both clusters contain seeds from both frameworks; with $n=9$ this is a suggestive cluster rather than an established threshold (a single seed in the empty band would dissolve it), but it indicates PR is a candidate framework-independent discriminator of expert play even though all 9 trainable seeds saturate FC1 at $\sim$64 active neurons.}
  \label{fig:ablation}
\end{figure}

\subsection{Within-Game Capacity-Limited Scaling at Matched Hyperparameters}
\label{sec:within-game}

Cross-game scaling (Section~\ref{sec:results}, Table~\ref{tab:main}) shows active neuron counts increasing with task complexity, but a skeptic could object that this depends on subjective state-variable estimates. The frozen rows of Table~\ref{tab:ablation} provide a stronger, hyperparameter-controlled test: \textit{within} a single game, with task and architecture and recipe held fixed, three independent random projections produce three different active-neuron counts (5, 7, 11) and three different rewards ($+14.1$, $+18.4$, $+18.7$). With $n=3$ this is suggestive rather than a fitted scaling law, and the meaningful split is binary---the 5-neuron seed plateaus below expert; the 7- and 11-neuron seeds reach indistinguishable expert reward---rather than smooth and monotonic. PR shows the same suggestive ordering: c120 at $9.63$, c20 at $11.74$, c10 at $13.73$.

The interpretation is direct. Each frozen CNN is a fixed random projection from $84\times84\times4$ pixels to a 3,136-dimensional feature space. The number of task-relevant directions present in that projection sets a ceiling on the FC1 active-neuron count: gradient descent cannot recruit a feature that is not there. Seeds c10, c20, c120 differ only in which random projection they happen to be, and that difference accounts for the bottleneck-size and performance differences we observe. Within-game evidence at matched recipe eliminates several interpretive degrees of freedom that cross-game evidence cannot (state-variable counts, reward-clipping conventions, action-space size); we report a directional finding rather than a slope, and additional same-recipe seeds would be needed to characterize the (active, reward) distribution.

\subsection{Effective Dimensionality and Bottleneck Signatures Across Regimes}
\label{sec:dimensionality}

The 6-seed Pong table reveals a second pattern in the trainable rows: t1 and t14 both reach expert reward at PR $\le 13$, while t2 stalls at $+14.4$ with PR $= 29.1$. Among 14 trainable Pong attempts under matched ablation-eval methodology, only 2 reached expert---trainable Pong is decidedly not a uniform high-success-rate regime under our recipe. The expert/non-expert split tracks PR: across 4 SB3 + 5 SF trainable seeds (Figure~\ref{fig:ablation}B; Appendix~\ref{app:cross_framework_pong}, Table~\ref{tab:cross_framework_pong}), all 4 with PR $\le 13$ reach expert; all 5 with PR $\ge 17$ do not, with no seed in the empty PR band $[12.6, 17.7]$. With $n=9$ this is a suggestive cluster rather than an established threshold---a single additional seed in the empty band would dissolve it---but PR provides a candidate discriminative signal that active count cannot, since FC1 saturates at 64 in all 9 cases. Both clusters contain seeds from both PPO frameworks, so the threshold is not a framework artifact.

Breakout shows a different pattern. Table~\ref{tab:breakout_signatures} contrasts three frozen training regimes against one trainable seed (b3 at 2000M) under sticky-action evaluation ($n_{\text{eval}}=100$ throughout). The first two frozen groups are deterministic-trained, evaluated at both $p_\text{sticky}=0$ (in-distribution) and $p_\text{sticky}=0.25$ (out-of-distribution); the third group is progressive-sticky-trained, evaluated at $p_\text{sticky}=0.25$ (in-distribution).

\begin{table}[ht]
\centering
\caption{SB3 Breakout cross-regime PR comparison under matched evaluation methodology ($n_{\text{eval}} = 100$, deterministic action selection; deterministic-trained sweep 2026-05-08, progressive-sticky sweep 2026-05-10). FC1 active counts are essentially regime-invariant (deterministic frozen: 20--24; progressive-sticky frozen: 17--25; trainable b3: 56/51, given as sticky=0/sticky=0.25).}
\label{tab:breakout_signatures}
\small
\setlength{\tabcolsep}{4pt}
\begin{tabular}{lcccc}
\toprule
 & \multicolumn{2}{c}{sticky=0 eval} & \multicolumn{2}{c}{sticky=0.25 eval} \\
\cmidrule(lr){2-3}\cmidrule(lr){4-5}
Run & PR & Reward & PR & Reward \\
\midrule
\multicolumn{5}{l}{\textit{Frozen, deterministic-trained (in-distribution at sticky=0)}} \\
seed 5 (400M)  & 9.37  & $221 \pm 16$ & 9.88  & $130 \pm 15$ \\
seed 10 (777M) & 7.68  & $351 \pm 3$  & 10.84 & $26 \pm 2$ \\
seed 0 (853M)  & 11.57 & $303 \pm 12$ & 9.90  & $26 \pm 1$ \\
\textit{mean}  & \textbf{9.54} & 292 & \textbf{10.21} & 61 \\
\midrule
\multicolumn{5}{l}{\textit{Frozen, progressive-sticky-trained (in-distribution at sticky=0.25)}} \\
progsticky\_c10 (1500M)         & -- & --  & 13.82 & $49 \pm 4$ \\
progsticky\_fast\_c10 (1416M)   & -- & --  & 16.66 & $63 \pm 6$ \\
progsticky\_continue\_c10 (2879M) & -- & --  & 12.57 & $53 \pm 5$ \\
\textit{mean}                    &    &     & \textbf{14.35} & 55 \\
\midrule
\multicolumn{5}{l}{\textit{Trainable (in-distribution at sticky=0.25)}} \\
b3 (2000M) & 8.94 & $388 \pm 5$ & \textbf{3.63} & $298 \pm 10$ \\
\midrule
\textbf{Frozen $-$ Trainable PR gap (sticky=0.25)} & & & $+6.58$ (det.) / $+10.72$ (prog.) & \\
\bottomrule
\end{tabular}
\end{table}

The matched-eval data refines the cross-regime claim along three readings of Table~\ref{tab:breakout_signatures}. \textit{At $p_\text{sticky}=0.25$ with deterministic-trained frozen agents evaluated OOD:} PR $\sim$10 vs trainable PR $= 3.63$, a $+6.6$-unit gap. \textit{At $p_\text{sticky}=0.25$ with progressive-sticky-trained frozen agents evaluated in-distribution:} PR $\sim$14 vs trainable PR $= 3.63$, a $+10.7$-unit gap---a stronger version of the structural-difference claim, with both policies functional at their training regime. \textit{At $p_\text{sticky}=0$ (deterministic-trained frozen, OOD trainable):} PR gap collapses to $+0.6$. Active count is the most robust cross-regime contrast, preserved across all rows (frozen 17--25 vs trainable 51--56).

The progressive-sticky-trained reading is the cleanest comparison because both policies are evaluated in-distribution. The deterministic-trained reading additionally serves as a robustness check that the frozen active set and PR survive eval-time perturbation even when the policy itself does not: $|\Delta\text{PR}|\le 3.2$ per frozen seed across eval conditions, despite $>90\%$ reward collapse for two of three deterministic seeds. \textbf{Active count and PR are complementary, not interchangeable}: active count answers ``which neurons participate?''; PR answers ``how independent are their contributions in the FC2 output?'' Both frozen regimes use a narrow set of moderately-diverse neurons (PR/FC1 $\sim$0.4--0.7); trainable uses a broad set of more-redundant neurons (PR/FC1 $\sim$0.16 at sticky=0). Characterizing a frozen-vs-trainable bottleneck difference requires reporting both metrics.

\section{The Bottleneck Locks Before Performance Emerges}
\label{sec:trajectory}

To probe when the bottleneck forms, we ran a 47-checkpoint longitudinal sweep across the three matched-recipe SB3 frozen Pong seeds c10, c20, c120 (15--16 checkpoints each, log-spaced from 5M to $\sim$1.5B steps), with full ablation analysis at every checkpoint. Two findings emerge cleanly.

\textbf{The full bottleneck signature is detectable from 5M steps.} Across all 47 ablations, FC1-Remove returns the random-play floor of $-21.0$. The active set's necessity does not develop over training: it is present from the earliest sampled checkpoint, before any reward improvement. FC1 active count itself locks by 15--30M steps in all three seeds, settling at 11 (c10), 7 (c20), and 5 (c120) with negligible drift over the subsequent $\sim$1.5B steps.

\textbf{Reward emerges 35--105M steps after the active-set lock, and the lag scales inversely with bottleneck capacity.} Table~\ref{tab:trajectory} summarizes the lock-vs-reward timing across seeds. The 11-neuron seed crosses reward $= 0$ within 35M steps after lock; the 7-neuron seed takes $\sim$65M; the 5-neuron seed takes $\sim$105M. Greater compression takes longer to learn to use, even though the compression itself crystallizes at the same point in training.

\begin{table}[ht]
\centering
\caption{Three SB3 frozen Pong seeds under matched progressive sticky recipe. The information bottleneck locks early (FC1 active count, FC1-Remove already at $-21.0$). Reward turns positive 35--105M steps later, with the lag inversely proportional to capacity. All values from the $n_{\text{eval}} = 100$ longitudinal sweep dated 2026-05-03; PR is FC2-output participation ratio at the final checkpoint.}
\label{tab:trajectory}
\small
\begin{tabular}{lcccccc}
\toprule
Seed & FC1 Active (locked) & Lock Step & Reward$=0$ Step & Lag & Final Reward & Final PR \\
\midrule
c10  & 11 & $\sim$15M & $\sim$50M  & $\sim$35M  & $+18.7$ & $13.31$ \\
c20  & 7  & $\sim$30M & $\sim$95M  & $\sim$65M  & $+18.5$ & $11.74$ \\
c120 & 5  & $\sim$30M & $\sim$135M & $\sim$105M & $+14.1$ & $8.38$  \\
\bottomrule
\end{tabular}
\end{table}

The strict timing of these events---FC1-Remove $\to -21.0$ at 5M; active-count lock at 15--30M; reward emergence at 50--135M---decisively rules out the alternative hypothesis that neurons die because the policy converged. The causality runs the opposite direction: representational compression is present in the very early phase, and downstream learning subsequently catches up. Full per-checkpoint data appear in Appendix~\ref{app:trajectory}.

\begin{figure}[ht]
  \centering
  \includegraphics[width=\linewidth]{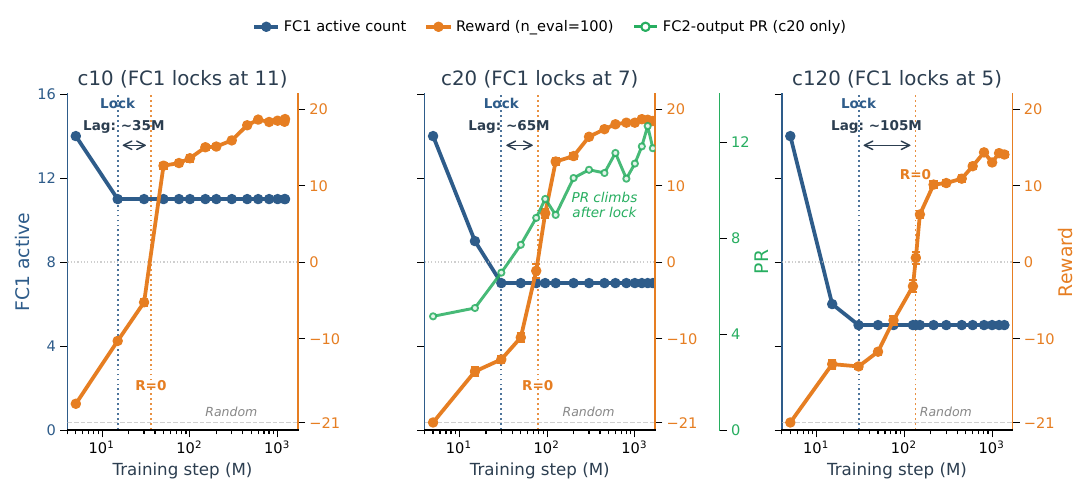}
  \caption{The information bottleneck locks before performance emerges, with lag inversely proportional to bottleneck capacity. Three same-recipe SB3 frozen Pong seeds (c10, c20, c120; left to right ordered by descending FC1 capacity 11/7/5). In each panel, the FC1 active count (blue, left axis) drops to its lock value within 15--30M steps and stays flat for the remaining 1.5B+ steps, while reward (orange, right axis) hovers near random play until 50--135M steps later. The vertical ``R$=$0'' marker uses the symmetric Pong reward midpoint as the lag threshold; lag values shift but the capacity-inverse-lag pattern persists under alternative thresholds. The lag widens as the bottleneck tightens---35M for c10, $\sim$65M for c20, $\sim$105M for c120---confirming that representational compression precedes functional competence. The c20 panel additionally shows that FC2-output PR (green; only plotted for c20 to keep the other panels readable) continues climbing for 1.5B+ steps after FC1 active count locks at 7, indicating a second learning process operating within an immutable feature subspace.}
  \label{fig:trajectory}
\end{figure}

The c20 seed makes a complementary point about post-lock learning: FC1 active count is locked at 7 from 30M onwards, but FC2-output participation ratio climbs from $\sim$6.6 at lock time to $\sim$11.7 at the 1.6B-step plateau, while reward climbs from $-12.7$ to $+18.5$. The 7 frozen-CNN feature channels do not change; the trainable FC head and policy/value heads continue to find richer ways to combine those channels.

\textbf{Reward emergence threshold.} ``Reward turns positive'' uses the symmetric Pong reward range midpoint (reward $= 0$). The qualitative conclusion---reward emerges much later than active-set lock---is robust to threshold choice, but specific lag values depend on it: at near-expert (reward $\ge +15$) the lags are $\sim$185/285/595M for c10/c20/c120, and at first-time-non-floor the lags compress to $\le 5$M for all seeds. The capacity-inverse-lag pattern persists across thresholds; its absolute scale does not.

\textbf{Active-set early commitment is also observed in trainable agents, with qualitatively different content.} Frozen networks have no choice---the random projection \textit{is} fixed---so an early lock in the frozen regime might appear trivial. Two longitudinal probes of trainable agents show otherwise. In trainable Breakout (b3, 17 checkpoints, 10M to 2B), the FC1 active set is committed by 10M: adjacent-checkpoint Jaccard overlap exceeds 0.90 throughout, with 47 of 61 union neurons present at both endpoints (Appendix~\ref{app:hump}). In stuck trainable Pong (t2, 10 checkpoints, 1M to 100M, Appendix~\ref{app:t2}), the active set saturates by 10M (44 $\to$ 64 neurons) and reward stabilizes by 20M, but FC2-output PR climbs through the full 100M without saturating. The two regimes commit early to qualitatively different things: frozen agents commit to \textit{which 5--26 neurons among 64 participate}, an informative sparse selection; trainable agents commit to \textit{participating broadly} (43--64 of 64), a weaker selection where the question becomes which few stay below threshold. Post-commitment trajectories differ across (game, regime, eval condition); we report early-stabilization signatures in both regimes without claiming a unified phenomenon.

\section{Discussion}
\label{sec:discussion}

\subsection{Why Sparsity Emerges: The Random Projection Mechanism}
\label{sec:mechanism}

Pong has approximately 5 real degrees of freedom, defining a 5-dimensional manifold in the $84 \times 84 \times 4 = 28{,}224$-dimensional pixel space. The CNN maps pixels to a 3,136-dimensional feature vector. In the trainable condition, the CNN learns to extract task-relevant dimensions; in the frozen condition, it applies a \textit{fixed random projection}. The story that follows---FC1 neurons aligning with task-relevant directions of the random projection---is a heuristic interpretation consistent with our empirical observations; the precise implicit bias driving emergent sparsity remains mathematically unproven (candidate mechanisms in Appendix~\ref{app:theory}). The Johnson-Lindenstrauss lemma guarantees that random projections approximately preserve distances on a low-dimensional manifold, provided the projection dimension exceeds intrinsic dimensionality. Each FC1 neuron computes $h_k = \text{ReLU}(\mathbf{w}_k^\top \phi(\mathbf{x}))$, a weighted combination of all 3,136 features, and through training each surviving neuron aligns its weight vector with a task-relevant direction---like tuning a radio antenna to a specific frequency. Remaining neurons find no independent task-relevant directions and become inert. When the CNN is trainable, each FC1 neuron sends gradient signal back through the CNN to manufacture custom features, and cooperative optimization fills as many independent directions as FC1 has capacity for ($\sim$60--64). The frozen CNN receives no gradients, so FC1 must work with whatever task-relevant directions exist in the random projection: for Pong, $\sim$1--3; for Breakout, 19--26; for Space Invaders, 42.

This cooperative dynamic in the trainable case explains the structural asymmetry observed in Breakout (Section~\ref{sec:dimensionality}): trainable agents activate 52--59 FC1 neurons but achieve participation ratio only $\sim$4 (under sticky=0.25 evaluation), indicating the activations are highly correlated. Two factors drive this redundancy. First, gradient descent on overparameterized networks distributes the same task signal across many neurons because each correlated path reduces variance on the policy gradient---a noise-averaging effect well-known in ensemble learning. Second, the CNN is free to manufacture features, so the same low-dimensional task-relevant signal can be replicated across many FC1 inputs at no representational cost. The frozen CNN cannot do this; if a feature direction is not present in the random projection, no number of FC1 neurons can synthesize it. Frozen agents are therefore forced to find the few mostly-orthogonal directions the projection happens to expose, yielding the narrow-and-diverse signature. The present work extends the principles of random projection and compressed sensing---where random projections of intrinsically low-dimensional signals naturally admit sparse representations---from static signal recovery to RL representation: gradient descent itself performs the sparse readout. A physical analogue from lensless imaging and full citation context are developed in Appendix~\ref{app:diffusercam}.

\paragraph{Freeway: a boundary case.}\label{sec:freeway} Freeway reveals the limits of frozen-CNN sparsity analysis. Its reward structure permits a degenerate ``always UP'' policy that scores $\sim$65\% of optimal without any visual processing, creating an overpowered local optimum that traps 17--25\% of trainable seeds as well, confirming it is a property of the game's reward landscape rather than a frozen-CNN artifact. Frozen Freeway agents in SB3 produce 1--5 active FC1 neurons, but ablation reveals a critical distinction: two runs show catastrophic FC1-Remove (reward drops to 0.0), confirming genuine information bottlenecks, while SF frozen runs and one SB3 seed produce entirely vestigial neurons where FC1-Remove has no effect. \textbf{Active neuron count alone is insufficient to diagnose information bottlenecks}; ablation verification is the ground truth. The frozen-CNN bottleneck reliably probes intrinsic task dimensionality only when the reward structure forces feature utilization, a precondition satisfied by the vast majority of RL environments. Full Freeway data appear in Appendix~\ref{app:freeway_complete}.

\paragraph{Practical implications and generality.} A frozen-CNN training run reveals, without explicit search, approximately how many independent state variables the agent's policy uses, informing curriculum design and pruning targets. Frozen CNNs also enable extreme-efficiency deployment: only the trained readout (e.g., $3 \times 3{,}136 \approx 9{,}400$ parameters for Pong) must be stored, with a random seed sufficient to regenerate the projection; frozen weights additionally cannot suffer representation drift or catastrophic forgetting. The core mechanism---a fixed random projection plus gradient descent on a sparse readout---requires nothing specific to RL, suggesting two broader applications that we offer as conjecture rather than result. \textit{First}, the same procedure may serve as a minimal-feature-set discovery tool wherever input dimensionality vastly exceeds task-relevant intrinsic dimensionality: a frozen random projection paired with a sparse-by-emergence readout could indicate the effective number of independent factors driving a downstream label, complementing techniques that require explicit sparsity penalties or post-hoc pruning. Domains with this structure include high-dimensional biological measurements (transcriptomic, proteomic, and metabolomic profiles where thousands of features predict a small number of phenotypic states) and image-based scientific classification more generally. \textit{Second}, the connection to LoRA \citep{aghajanyan2020intrinsic, hu2021lora} is more than analogical: LoRA's success rests on the assumption that task adaptation lives in a low-dimensional subspace of the pretrained model's parameter space, and our results suggest the readout layer of a frozen feature extractor will spontaneously reveal that subspace's effective rank without requiring it to be specified in advance---a lightweight diagnostic for adaptation-rank selection. Both conjectures follow the same mechanism that produces the Atari results, and we offer them as concrete starting points for extending the present work beyond reinforcement learning.

\paragraph{Sticky actions as a complexity-modifying intervention.} Sticky actions are most often introduced as a methodological safeguard against memorized action sequences. Our results suggest a complementary interpretation: sticky actions modify the effective task complexity. Each agent must now compensate for occasional unintended action repetition, expanding the set of state-action contexts the policy must distinguish. Consistent with our central thesis, this expansion shows up in our data---within Pong alone (same game, same architecture, same recipe up to the curriculum), the frozen FC1 active count rises from 1--3 (deterministic) to 5--11 (progressive sticky reaching $p_\text{sticky}=0.25$). The bottleneck adjusts to match. This is a within-game replication of the cross-game scaling effect, achieved by adding stochasticity rather than changing games, and it indicates that ``task complexity'' as the bottleneck reflects it includes the dynamics regime, not just the static structure of state space.

\subsection{Limitations}
\label{sec:limitations}

\textbf{Game and architecture coverage.} Four games span a complexity range but are insufficient to fit a quantitative scaling law; results use the Nature-DQN CNN, and different architectures (ResNets, Vision Transformers) may produce different sparsity patterns. All games studied are discrete-action domains; whether emergent sparsity transfers to continuous-control settings (e.g., MuJoCo locomotion or robotic manipulation) where the action space geometry is fundamentally different remains an open question for future work. We lack a formal proof for when frozen random projections guarantee sparse gradient-descent solutions; candidate mechanisms appear in Appendix~\ref{app:theory}.

\textbf{Sticky actions and seed lottery.} Pong and Breakout have inherently reactive dynamics that mitigate memorization concerns \citep{machado2018revisiting}; nonetheless, fixed $p_\text{sticky}=0.25$ training of frozen Breakout did not learn under our V7 setup. A reward-paced progressive sticky curriculum rescued the seed-10 lineage to competent (reward $\sim$50--65) play at $p_\text{sticky}=0.25$ evaluation, though not to expert (reward $\geq 200$). The progressive curriculum was similarly load-bearing for stochastic frozen Pong. Cross-framework trainable success rates differ substantially (Appendix~\ref{app:crossframework}); the PR cluster in Figure~\ref{fig:ablation}B contains seeds from both frameworks in both clusters, indicating the PR-vs-expert separation is not a framework artifact.

\section{Conclusion}

We have demonstrated that sparsity emerges spontaneously in deep RL agents when CNN feature extractors are frozen at random initialization. Without any explicit sparsity objective, gradient descent discovers solutions using 1--11 neurons for Pong, 19--26 for Breakout, and $\sim$42 for Space Invaders, while trainable CNNs use 55--64 regardless of task complexity. Cross-game scaling tracks task complexity. Within-game scaling at matched recipe produces 5/7/11 active $\to +14$/$+18$/$+19$ reward across three same-recipe seeds---a directional finding rather than a fitted scaling law---consistent with the random projection's usable dimensionality setting a ceiling on performance even at fixed task and capacity. Ablation confirms necessity across both PPO frameworks. A 47-checkpoint longitudinal sweep shows the bottleneck signature is present from 5M and the active set locks by 15--30M, while reward turns positive 35--105M steps later (using reward $= 0$), with the lag widening as bottleneck capacity tightens across the three seeds. Active-set early stabilization recurs in trainable agents but with qualitatively different content---a broad set rather than a narrow one.

A separate cross-regime finding bears emphasis: in Breakout, frozen and trainable CNNs reach competitive reward via structurally different bottleneck signatures. Under matched $p_\text{sticky}=0.25$ evaluation, frozen agents use 17--25 active neurons at PR $\sim$10--14 while trainable agents use 51 at PR $\sim$3.6. Active count and participation ratio are complementary, not interchangeable, indicators of compression. Together these results connect lottery ticket sparsity, reservoir random features, and compressed sensing through one observable phenomenon: the frozen random CNN, acting as a compressive measurement device, reveals the intrinsic dimensionality that trainable networks obscure.

\bibliography{references_v30}
\bibliographystyle{tmlr}

\appendix

\section{Width Invariance}
\label{app:width}

A natural objection is that observed sparsity reflects a fixed fraction of network capacity. We address this with width-scaling experiments on Pong (SB3, frozen CNN).

\begin{table}[ht]
\centering
\caption{Width-scaling results for Pong frozen CNNs. Active counts do not scale proportionally with width.}
\label{tab:width}
\small
\begin{tabular}{llcccc}
\toprule
FC1 Width & Seed & Active & \% Used & Max Reward \\
\midrule
32 & Seed 4 & 5 & 15.6\% & +19.1 \\
\midrule
\multirow{3}{*}{64} & Seed 2 & 3 & 4.7\% & +19.2 \\
 & Seed 4 & 6 & 9.4\% & +19.8 \\
 & Seed 3 & 8 & 12.5\% & +20.7 \\
\midrule
\multirow{2}{*}{128} & Seed 43 & 5 & 3.9\% & +20.7 \\
 & Seed 42 & 14 & 10.9\% & +19.2 \\
\bottomrule
\end{tabular}
\end{table}

Under a capacity-proportional model, predicted active counts would scale linearly (e.g., 3.2, 6.4, 12.8 at 10\% utilization). Instead, ranges overlap substantially: a 128-width network solves Pong with 5 neurons (matching the 32-width result), and within 128-width alone, seeds produce 5 and 14 active neurons, a 2.8$\times$ ratio comparable to the entire cross-width range. Seed variance exceeds width effects, and percentage utilization varies inversely with width (3.9--15.6\%), confirming that task structure, not capacity, governs solution size.

\section{Cross-Framework Validation and Initialization Sensitivity}
\label{app:crossframework}

The frozen-CNN sparsity phenomenon replicates across two independent PPO implementations differing in synchronicity (SB3 synchronous vs.\ SF asynchronous), hardware (Mac vs.\ Ubuntu/RTX~5090), and software stack. Both produce FC1 bottlenecks of 1--3 neurons for Pong with identical ablation patterns (FC1-Remove $= -21.0$). Both the SB3 ablation run ($p_\text{sticky} = 0.25$ via wrapper) and all SF runs ($p_\text{sticky} = 0.25$ at the ALE level) include sticky actions.

Cross-framework replication succeeded for Pong but not for Breakout. Extensive investigation (14 identified differences systematically eliminated across 9 SF configuration versions) revealed that SF's asynchronous PPO (APPO) introduces policy lag that prevents frozen-CNN Breakout agents from learning, while trainable SF Breakout achieves competitive reward ($\sim$143 at 200M). Frozen CNNs appear more sensitive to policy staleness because the fixed feature space offers no room for the CNN to adapt its representations to compensate for stale gradients.

\paragraph{Trainable Pong success-rate asymmetry between frameworks.} Under matched ablation-eval methodology with $n_{\text{eval}} = 100$, only 2 of 14 SB3 trainable Pong seeds reached expert ($\sim$14\%); under SF V5 (fixed sticky, asynchronous PPO) the rate was 5 of 5 trainable. We do not have a clean explanation for this gap. Two candidates: (i) SF's APPO policy lag may regularize toward smoother solutions on simpler tasks; and (ii) SF's V5 default policy-head initialization (Kaiming uniform with large initial logits, see below) produces an entropy collapse from initialization that may paradoxically push trainable agents through the early Pong learning regime more reliably than SB3's well-conditioned start. A controlled cross-framework ablation isolating these factors is left for future work. A related observation: the progressive sticky curriculum was load-bearing for stochastic frozen Pong (3/$\sim$20 SB3 seeds passed a 10M-step screening criterion under the curriculum; 0/15 SF frozen Pong seeds passed under fixed $p_\text{sticky}=0.25$). The PR cluster in Figure~\ref{fig:ablation}B contains seeds from both frameworks in both expert and stuck clusters, indicating the PR-vs-expert separation is not a framework artifact even though the absolute success rates are.

\subsection*{Policy Head Initialization: The V5/V6 Finding}

During SF experiments, we discovered that policy head initialization dramatically affects early training dynamics. SF's default Kaiming initialization produces initial logits of magnitude 28--55, driving entropy to $\sim$0.0 within 500K steps. Switching to orthogonal initialization with std$=$0.01 (matching SB3's default) produces initial logits of 0.4--1.4, eliminating entropy collapse entirely.

\begin{table}[ht]
\centering
\caption{SF Pong frozen seed success rates under Kaiming (V5) and orthogonal (V6) policy head initialization. The $\sim$25\% success rate is stable; which seeds learn is shuffled.}
\label{tab:v5v6}
\small
\begin{tabular}{lccc}
\toprule
Init & Learners / Total & Success Rate & Entropy Collapse? \\
\midrule
Kaiming & 4/16 & 25\% & Yes (universal) \\
Orthogonal & 4/15 & 27\% & No \\
Combined & 8/15 (unique) & 53\% & --- \\
\bottomrule
\end{tabular}
\end{table}

The two initialization conditions have \textbf{zero overlap}: no CNN seed that learns under one also learns under the other (out of 15 shared seeds). Yet 8 of 15 seeds (53\%) learn under at least one. This reveals that $\sim$25\% of random projections support learning per initialization condition, but \textit{which} projections succeed depends on the CNN geometry $\times$ optimization trajectory interaction. Crucially, both conditions produce the same 1--2 neuron bottleneck with identical ablation patterns; the sparsity is a property of frozen-feature learning itself, not any particular initialization.

\section{Complete Pong Results}
\label{app:pong_complete}

\textbf{Pong (deterministic, 200M).} Across five SB3 frozen seeds, FC1 active neuron counts range 3--8, all achieving expert-level play (+16 to +20 reward) with only 5--12\% of available FC1 neurons. An extended SB3 run (800M steps under $p_\text{sticky} = 0.25$) converged to 3 active neurons at +17.8 reward. SF frozen agents independently converged to 1--2 active neurons across five seeds, achieving lower absolute reward (reflecting the harder optimization landscape of asynchronous PPO on this task) but the same extreme bottleneck.

\textbf{Breakout (no sticky).} Four frozen seeds on Mac/Ubuntu SB3 PPO, FC1 active 19--26 (mean $\approx 23$). Extended training with a continuation learning-rate schedule (warmup 0$\to 5 \times 10^{-5}$ over 10M, then linear decay) pushed our best seed to 355 reward (deterministic), approaching the $\sim$400 trainable baseline. The 19--26 spread is broader than Space Invaders' uniformly 42, suggesting moderately complex tasks admit more variation in how gradient descent partitions information across the random feature basis.

\textbf{Breakout (progressive sticky).} Three frozen seed-10-lineage SB3 PPO runs trained under a reward-paced progressive sticky curriculum ($p_\text{sticky}$ ramping from 0 to 0.25 as training reward approaches 200) and evaluated at $p_\text{sticky}=0.25$ ($n_{\text{eval}}=100$, sweep 2026-05-10): FC1 active 17--25 (mean 20), reward 49--63 (mean 55), PR 12.57--16.66 (mean 14.35), FC1-Remove crashes to $\sim$2--3 in all three. Fixed $p_\text{sticky}=0.25$ training of seed 10 originally stalled at reward $\sim$3 under our V7 SF setup; the progressive curriculum recovered the seed-10 lineage to competent (but not expert) play. Notably, the lucky deterministic-trained seed 5 evaluated OOD at $p_\text{sticky}=0.25$ reached reward 130 (Appendix~\ref{app:freeway_complete}), exceeding any progressive-sticky-trained run; progressive sticky lifts the floor of frozen Breakout sticky=0.25 performance but does not match the best generalization-lucky deterministic seed.

\textbf{Space Invaders (sticky 0.25).} Two frozen seeds, both converging to exactly 42 active FC1 neurons.

\begin{table}[ht]
\centering
\caption{SB3 Pong frozen-CNN results across seeds (200M steps, no sticky actions). Active neuron counts range 3--8 while trainable controls use 58--64.}
\label{tab:pong_sb3}
\small
\begin{tabular}{lccccc}
\toprule
Experiment & FC1 Active & Max Reward & Final Reward & Steps & Cross@0.0 \\
\midrule
Seed 2 (64$\times$2) & \textbf{3} & +19.2 & +18.8 & 200M & 70.5M \\
Seed 1 (64$\times$1) & 4 & +18.7 & +16.3 & 200M & 124.5M \\
Seed 4 (64$\times$2) & 6 & +19.8 & +19.5 & 200M & 103.7M \\
Recovery (64$\times$64) & 8 & +20.1 & +19.8 & 100M & 72.8M \\
Seed 3 (64$\times$2) & 8 & +20.7 & +20.1 & 137M & 44.6M \\
\midrule
Trainable (5 seeds) & 58--64 & +20.9 & +20.7 & 10M & $<$5M \\
\bottomrule
\end{tabular}
\end{table}

\begin{table}[ht]
\centering
\caption{Complete Pong ablation results across all conditions and frameworks.}
\label{tab:pong_complete}
\small
\begin{tabular}{llccccccc}
\toprule
FW & Model & FC1 & FC2 out & PCA95 & Baseline & FC1-Keep & FC1-Rem & FC2-Rem \\
\midrule
\multicolumn{9}{l}{\textit{SB3 Frozen (800M, sticky $p=0.25$)}} \\
SB3 & Frozen seed 1 & 3 & 46 & 3 & +15.2 & +14.8 & $-$21.0 & $-$21.0 \\
\midrule
\multicolumn{9}{l}{\textit{SF Frozen (200--800M, sticky $p=0.25$, Kaiming head)}} \\
SF & Frozen seed A & 1 & 9 & 1 & $-$0.9 & $-$2.0 & $-$21.0 & $-$21.0 \\
SF & Frozen seed B & 2 & 29 & 2 & $-$3.8 & $-$4.3 & $-$21.0 & $-$21.0 \\
SF & Frozen seed C & 2 & 34 & 2 & $-$5.2 & $-$5.5 & $-$21.0 & $-$21.0 \\
\midrule
\multicolumn{9}{l}{\textit{SF Frozen (199--800M, sticky $p=0.25$, Orthogonal head)}} \\
SF & Frozen seed D & 1 & 12 & 1 & $-$11.5 & $-$10.9 & $-$21.0 & $-$21.0 \\
SF & Frozen seed E & 1 & 10 & 1 & $-$11.2 & $-$11.6 & $-$21.0 & $-$21.0 \\
\midrule
\multicolumn{9}{l}{\textit{Trainable Controls}} \\
SF & Trainable & 64 & 60 & 52 & +19.7 & +19.6 & $-$21.0 & $-$21.0 \\
\midrule
\multicolumn{9}{l}{\textit{Dead Seed (Gradient Collapse)}} \\
SF & Dead seed & 0 & 0 & --- & $-$21.0 & $-$21.0 & $-$21.0 & $-$21.0 \\
\bottomrule
\end{tabular}
\end{table}

\section{Mechanistic Analysis of the 3-Neuron Pong Solution}
\label{app:mechanistic}

The SB3 Pong frozen Seed~2 (200M steps) achieves expert performance (+19.2 max reward) with exactly 3 active FC1 neurons: indices 42, 47, and 54.

\begin{table}[ht]
\centering
\caption{Fisher discriminant ratios and value correlations for the 3 active neurons. Neuron~54 has near-zero action discrimination but strong value encoding.}
\label{tab:fisher}
\small
\begin{tabular}{lccc}
\toprule
Neuron & Fisher Ratio & Corr.\ with $V$ & Linear $R^2$ (value) \\
\midrule
N42 & 0.724 & 0.513 & 0.264 \\
N47 & 0.801 & 0.277 & 0.077 \\
N54 & \textbf{0.014} & \textbf{0.516} & \textbf{0.267} \\
\midrule
All 3 (MLP) & --- & --- & \textbf{0.993} \\
\bottomrule
\end{tabular}
\end{table}

Neuron~54's Fisher ratio of 0.014 means it provides almost no information for discriminating between UP, DOWN, and NOOP actions. Yet ablating it drops performance from +19.4 to +7.4, a catastrophic 12-point decline. This resolves when we observe its strong value correlation: neuron~54 is a \textit{value encoder}, not an action discriminator. The PPO architecture uses shared features for both policy and value heads, so the 3-neuron bottleneck must jointly serve both functions. Neurons~42 and 47 primarily support action selection; neuron~54 primarily supports value estimation; together they achieve $R^2 = 0.993$ for value prediction.

The three neurons support action discrimination among effectively 3 actions (UP, DOWN, NOOP). Action centroids in the 3D activation space show clear spatial structure: UP$\leftrightarrow$DOWN centroid distance is 127.3, UP$\leftrightarrow$NOOP is 107.2, but DOWN$\leftrightarrow$NOOP is only 20.5, consistent with Pong's asymmetric control law where ``move up'' is well-separated while ``move down or stay'' are nearly aliased.

\section{Complete Freeway Results and Cross-Game Sticky-Robustness Check}
\label{app:freeway_complete}

\paragraph{Frozen Breakout cross-regime robustness check (matched $2\times 2$).} The cross-regime PR comparison reported in Section~\ref{sec:dimensionality} is supported by a controlled $2\times 2$ ablation sweep at $n_{\text{eval}}=100$, sweep dated 2026-05-08. We re-evaluated all three frozen Breakout headline seeds (seed5 @ 400M, seed10 @ 777M, seed0 @ 853M) at the alternate sticky condition ($p_\text{sticky}=0.25$, the trainable's training regime). Three observations.

\textit{First}, frozen FC1 active count is essentially regime-invariant (per-seed differences of $\le 4$ neurons, mean active count 22.0 at sticky=0 vs 21.7 at sticky=0.25). \textit{Second}, frozen PR is reasonably stable across regimes (per-seed $|\Delta\text{PR}|\le 3.16$, average +0.67), so the matched-eval PR comparison reported in Table~\ref{tab:breakout_signatures} differs from the original mismatched comparison only modestly. \textit{Third}, frozen \textit{reward} is highly fragile to sticky perturbation: seed10 drops from 350.6 to 25.6 and seed0 from 302.8 to 26.1 (both $>90\%$ reward collapse) when evaluated under sticky=0.25 they were not trained for. The active-neuron set carries the same identity, but the policy that reads from those neurons is calibrated to deterministic dynamics and breaks under sticky perturbation.

The robustness check in earlier drafts---a single deterministic-trained frozen Breakout seed showing FC1-Remove $\to 2.0$ under sticky evaluation---demonstrated only that the active set continues to be \textit{necessary} under regime perturbation; it did not establish that the policy reading from those neurons remains \textit{functional}. The $2\times 2$ data clarifies that those are different statements: the representation (active set, PR) is regime-stable, but the policy is regime-fragile.

\paragraph{Frozen Breakout progressive-sticky in-distribution evaluation (third group in Table~\ref{tab:breakout_signatures}).} A follow-up sweep (2026-05-10, $n_{\text{eval}}=100$, $p_\text{sticky}=0.25$) evaluated three progressive-sticky-trained Breakout runs in-distribution: progsticky\_c10 @ 1500M (FC1=17, PR=13.82, reward $48.9 \pm 4.4$), progsticky\_fast\_c10 @ 1416M (FC1=25, PR=16.66, reward $62.6 \pm 5.8$), and progsticky\_continue\_c10 @ 2879M (FC1=19, PR=12.57, reward $53.3 \pm 4.7$). All three reach competent (50--63) but not expert ($\geq 200$) play, with FC1-Remove $\to 2$--$3$ in all three (active-set necessity preserved). Mean active count 20, mean PR 14.35---PR $\sim$4 units higher than the OOD-evaluated deterministic-trained seeds, consistent with policies trained on stochastic dynamics using more diverse output combinations. The combined frozen-Breakout-at-sticky=0.25 picture across both training regimes (6 runs total): active 17--25, PR 9.88--16.66, reward 26--130 with the high tail being seed5's lucky OOD generalization and the typical case being reward 50--65 under in-distribution evaluation. Trainable b3 at sticky=0.25 reaches expert reward (298) at PR=3.63, giving the cross-regime structural-difference claim a $+6.6$ to $+10.7$-unit PR gap depending on which frozen training regime is read.

\paragraph{Freeway full ablation table.} Below.

\begin{table}[ht]
\centering
\caption{Complete Freeway results. ``Mech.'' indicates whether ablation confirms functional (F) or vestigial (V) neurons. SB3 trainable seeds marked $\star$ are collapsed ``always UP'' policies.}
\label{tab:freeway_complete}
\small
\begin{tabular}{lllccccl}
\toprule
FW & Condition & Model & FC1 & Reward & FC1-Rem & FC2-Rem & Mech. \\
\midrule
\multicolumn{8}{l}{\textit{Frozen --- SB3}} \\
SB3 & no sticky & Seed 1 & 4 & 22.0 & 21.9 & 21.9 & V \\
SB3 & no sticky & Seed 4 & 5 & 20.8 & 0.0 & 22.4 & F \\
SB3 & sticky 0.25 & Seed 1 & 2 & 21.2 & 0.0 & 0.0 & F \\
\midrule
\multicolumn{8}{l}{\textit{Frozen --- SF (all vestigial)}} \\
SF & sticky 0.25 & Seed A & 0 & 21.1 & 21.5 & 21.4 & V \\
SF & sticky 0.25 & Seed B & 2 & 22.7 & 21.1 & 21.3 & V \\
SF & sticky 0.25 & Seed C & 1 & 22.2 & 21.5 & 20.8 & V \\
\midrule
\multicolumn{8}{l}{\textit{Trainable --- SB3 (non-collapsed)}} \\
SB3 & no sticky & Seed 7 & 57 & 22.5 & 21.9 & 0.0 & F \\
SB3 & sticky 0.25 & Seed 6 & 62 & 30--31 & 0.0 & 0.0 & F \\
\midrule
\multicolumn{8}{l}{\textit{Trainable --- SB3 (collapsed ``always UP'')$\star$}} \\
SB3 & no sticky & Seed 1$\star$ & 12 & 21.8 & 21.9 & 21.9 & V \\
SB3 & sticky 0.25 & Seed 3$\star$ & 16 & 21.9 & 22.0 & 21.9 & V \\
\midrule
\multicolumn{8}{l}{\textit{Trainable --- SF (10 seeds, all expert $\sim$33.5)}} \\
SF & sticky 0.25 & 10 seeds & 22--56 & 33.3--33.8 & 0--21.5 & --- & F \\
\bottomrule
\end{tabular}
\end{table}

Key observations: SF frozen seed A achieves reward 21.1 with \textit{zero} active neurons in both FC1 and FC2 output; the entire policy is encoded in bias terms, confirming the ``always UP'' attractor. SB3 frozen seeds under sticky actions and without sticky actions include both functional and vestigial bottlenecks, demonstrating that ablation verification is essential. SB3 trainable seeds 1 and 3 collapse to ``always UP'' with vestigial neurons, confirming the attractor is not frozen-CNN-specific. All 10 SF trainable seeds reach expert-level reward (33.3--33.8) with zero collapses, likely due to orthogonal head initialization maintaining high initial entropy.

\section{Complete Longitudinal Data}
\label{app:trajectory}

\begin{table}[ht]
\centering
\caption{47-checkpoint longitudinal sweep across SB3 frozen Pong seeds c10, c20, c120 (15--16 checkpoints each, log-spaced 5M to $\sim$1.6B steps, $n_{\text{eval}} = 100$). All values from the 2026-05-03 sweep. FC1 active count is tightly stable post-lock; FC2-output participation ratio (PR) continues to climb. The FC1-Remove column is omitted because it equals $-21.0$ at every checkpoint in every seed (the bottleneck signature is present from 5M onward).}
\label{tab:trajectory_full}
\footnotesize
\setlength{\tabcolsep}{4pt}
\begin{tabular}{rccc|rccc|rccc}
\toprule
\multicolumn{4}{c}{\textit{Seed c10 (locks at 11)}} & \multicolumn{4}{c}{\textit{Seed c20 (locks at 7)}} & \multicolumn{4}{c}{\textit{Seed c120 (locks at 5)}} \\
Step(M) & FC1 & PR & R & Step(M) & FC1 & PR & R & Step(M) & FC1 & PR & R \\
\midrule
5    & 14 & 5.14  & $-$18.55 & 5    & 14 & 4.74  & $-$21.00 & 5    & 14 & 4.58 & $-$21.00 \\
15   & 11 & 5.74  & $-$10.32 & 15   & 9  & 5.09  & $-$14.32 & 15   & 6  & 5.55 & $-$13.36 \\
30   & 11 & 9.10  & $-$5.26  & 30   & 7  & 6.56  & $-$12.74 & 30   & 5  & 5.86 & $-$13.66 \\
50   & 11 & 9.82  & $+$12.63 & 75   & 7  & 8.84  & $-$1.14  & 75   & 5  & 6.60 & $-$7.63  \\
100  & 11 & 11.74 & $+$13.60 & 95   & 7  & 9.63  & $+$6.40  & 125  & 5  & 7.39 & $-$3.16  \\
200  & 11 & 12.98 & $+$15.13 & 125  & 7  & 8.97  & $+$13.17 & 135  & 5  & 7.10 & $+$0.56  \\
450  & 11 & 14.77 & $+$17.92 & 300  & 7  & 10.85 & $+$16.41 & 215  & 5  & 8.28 & $+$10.15 \\
800  & 11 & 14.14 & $+$18.35 & 800  & 7  & 10.48 & $+$18.26 & 600  & 5  & 9.21 & $+$12.56 \\
1215 & 11 & \textbf{13.31} & $+$\textbf{18.73} & 1601 & 7 & \textbf{11.74} & $+$\textbf{18.48} & 1378 & 5 & \textbf{8.38} & $+$\textbf{14.09} \\
\bottomrule
\end{tabular}
\end{table}

The complete 47-row CSV (all checkpoints for all three seeds, including FC1-Keep, FC2-output ablation, and PCA95) is available in the supplementary data.

\section{Cross-Framework Pong Trainable PR Picture}
\label{app:cross_framework_pong}

Across nine trainable Pong seeds (4 SB3 + 5 SF V5) under matched-eval methodology at $n_{\text{eval}} = 100$, FC2-output participation ratio cleanly separates expert from non-expert outcomes (Figure~\ref{fig:ablation}B).

\begin{table}[ht]
\centering
\caption{Cross-framework trainable Pong, all values at $n_{\text{eval}} = 100$, sticky = 0.25, deterministic eval, sweep dated 2026-05-02. SB3 seeds use progressive sticky and ent\_coef = 0.02; SF V5 seeds use fixed sticky = 0.25 and ent\_coef = 0.01. The two SF expert seeds (c20, c40) were captured at very early checkpoints (8M, 5M) before FC1 saturated to 64. The empty PR band $[12.59, 17.71]$ contains no seed from either framework.}
\label{tab:cross_framework_pong}
\small
\begin{tabular}{lcccc}
\toprule
Seed & Framework & FC1 active & \textbf{PR} & Reward \\
\midrule
\multicolumn{5}{l}{\textit{Expert (PR $\le 13$, reward $\ge +18$)}} \\
c40    & SF V5 & 36 (early ckpt) & 9.33  & $+18.29$ \\
t14    & SB3   & 64              & 11.37 & $+20.57$ \\
c20    & SF V5 & 36 (early ckpt) & 11.80 & $+19.01$ \\
t1     & SB3   & 64              & 12.59 & $+20.86$ \\
\midrule
\multicolumn{5}{l}{\textit{Stuck (PR $\ge 17$, reward $\le +15$)}} \\
t3     & SB3   & 64 & 17.71 & $+14.76$ \\
c50    & SF V5 & 64 & 19.13 & $+7.76$  \\
c10    & SF V5 & 63 & 21.11 & $+11.89$ \\
c30    & SF V5 & 64 & 22.42 & $+11.50$ \\
t2     & SB3   & 64 & 29.14 & $+14.52$ \\
\bottomrule
\end{tabular}
\end{table}

All 4 trainable seeds with PR $\le 13$ reach expert ($+18$ to $+21$); all 5 with PR $\ge 17$ do not. We report this as a Pong-specific empirical observation. Two caveats are worth flagging. First, the two SF expert seeds were captured at very early checkpoints (5M, 8M) before FC1 saturated to 64---their ``expert PR'' may partly reflect early training dynamics rather than a converged trainable bottleneck; nonetheless, the two SB3 expert seeds (t1, t14) at fully-saturated FC1$=64$ also satisfy PR $\le 13$, so the threshold is not an artifact of FC1 saturation state. Second, comparing this $n=100$ table against earlier $n=30$ readings, PR estimates were largely stable ($|\Delta\text{PR}| \le 1.5$ for all but two seeds) but stuck SF V5 \textit{reward} values shifted upward substantially at $n=100$ (e.g., c10 reward $+7.3 \to +11.89$, c30 $+2.8 \to +11.50$): $n=30$ underestimated stuck-policy reward magnitude due to high episode-to-episode variance, while PR remained the more reliable cluster discriminator. The PR-vs-expert separation replicates across two PPO implementations, two sticky regimes (progressive vs.\ fixed), and two entropy coefficients.

\section{$n_{\text{eval}}$ Calibration}
\label{app:n_eval}

Single-checkpoint PR estimates carry sample-noise variance that depends on policy stochasticity at evaluation. We re-ran all 11 manuscript headline ablations at both $n_{\text{eval}} = 30$ and $n_{\text{eval}} = 100$ to characterize this.

\begin{table}[ht]
\centering
\caption{Headline ablations at $n_{\text{eval}} = 30$ vs.\ $100$. Pong PR estimates are stable across $n$ (all $|\Delta\text{PR}| \le 2.1$); Breakout frozen PR estimates shift upward systematically; trainable Breakout b3 at 2000M shows a large $n=30$ outlier that resolves at $n=100$. Two independent $n=100$ Pong sweeps (Apr 18--20 vs.\ May 2) further differ by up to $\pm 2$ PR units for individual trainable seeds---we report the May 2 values throughout the manuscript.}
\label{tab:n_eval}
\small
\begin{tabular}{llcccc}
\toprule
Seed & Game/Regime & Reward (n=100) & PR (n=30) & PR (n=100) & $\Delta$PR \\
\midrule
c120 & Pong / frozen & $+14.1$ & 9.11  & 9.63  & $+0.52$ \\
c20  & Pong / frozen & $+18.4$ & 12.21 & 11.74 & $-0.47$ \\
c10  & Pong / frozen & $+18.7$ & 14.45 & 13.73 & $-0.72$ \\
t2   & Pong / trainable (stuck) & $+14.4$ & 30.33 & 29.14 & $-1.19$ \\
t1   & Pong / trainable & $+20.5$ & 12.71 & 12.59 & $-0.12$ \\
t14  & Pong / trainable & $+20.6$ & 11.50 & 11.37 & $-0.13$ \\
\midrule
seed 5  & Breakout / frozen & 221.1 & 6.46 & 9.37  & $+2.91$ \\
seed 0  & Breakout / frozen & 302.8 & 8.98 & 11.57 & $+2.59$ \\
seed 10 & Breakout / frozen & 350.6 & 2.95 & 7.68  & $+4.73$ \\
b2 & Breakout / trainable & 142.2 & 3.85 & 3.67 & $-0.18$ \\
b3 (2000M) & Breakout / trainable & 298.0 & 9.62 & 3.63 & $-5.99$ \\
\bottomrule
\end{tabular}
\end{table}

The pattern is informative. Pong PR is stable across the noise levels (all $|\Delta\text{PR}| \le 2.1$). Breakout frozen PR systematically rises by 2.6--4.7 units at $n=100$ (a regime where high-reward play has more diverse policy decisions, so larger samples reveal more variance directions). Trainable Breakout b3 has a single $n=30$ outlier at 2000M that resolves cleanly at $n=100$. Within-table orderings (e.g., c120 $<$ c20 $<$ c10 in frozen Pong) are preserved across $n$. We additionally observed that two independent $n=100$ sweeps run on different days (Apr 18--20 vs.\ May 2 for the Pong rows above) can differ by up to $\pm 2$ PR units for individual trainable seeds: this reflects residual eval-time stochasticity and informs the choice to use $n=100$ throughout the manuscript while reading single-decimal precision as approximate.

\section{The Hump and Eval-Condition Sensitivity in Trainable Breakout}
\label{app:hump}

A 17-checkpoint longitudinal sweep of trainable Breakout seed b3 (10M to 2000M) at $n_{\text{eval}} = 30$, $p_\text{sticky}=0.25$, showed a hump-shaped FC2-output PR trajectory: rising from $\sim$3.5 at 10M to $\sim$9 at 168M, then falling back to $\sim$4 by 2000M. We do not promote this to a primary finding because:
(i) at $n_{\text{eval}} = 100$ the hump weakens (peak at 168M drops only modestly from 9.04 to 9.42, but the late-trajectory descent partially attenuates: at the 2000M endpoint PR is 3.63 at $n=100$ but 8.94 when re-evaluated at sticky $= 0$);
(ii) a parallel longitudinal sweep on b2 (11 checkpoints, 10M to 1200M, $n=100$) shows a flat PR trajectory in the 2.9--4.2 band throughout---no hump. The hump is $n = 1$;
(iii) the post-peak descent under sticky $= 0.25$ partly reflects the policy learning to suppress sticky-induced exploration; under sticky $= 0$ the same checkpoint reveals a broader representation, so the trajectory shape is partly an eval-condition artifact rather than a pure representation-level property.

Frozen Breakout seed 10 by contrast shows plateau/oscillating PR (11--13 band) across 800M training steps at sticky $= 0$. Three trajectory archetypes appear in our data (frozen Pong: monotonic; frozen Breakout: plateau; stuck trainable Pong t2: gradual climb without saturation), and trainable Breakout b3 occupies a fourth shape category that is partly eval-condition-mediated. Full per-checkpoint b3 and b2 longitudinal data, plus the b3 sticky $= 0$ control, are available in supplementary CSVs.

\section{t2 Longitudinal: PR Climb in Stuck Trainable Pong}
\label{app:t2}

The Pong trainable t2 seed reaches reward $+14.76$ at 100M with FC2-output PR $= 31.26$. Whether the high-PR signature was committed early (analogous to b3's Jaccard finding) or evolved into via gradual diversification was tested with a 10-checkpoint longitudinal sweep (1M to 100M, $n_{\text{eval}} = 100$, sticky = 0.25).

\begin{table}[ht]
\centering
\caption{t2 longitudinal trajectory. Active set saturates by 10M steps and reward stabilizes by 20M, but FC2-output PR continues monotonically climbing for the remaining 80M.}
\label{tab:t2_longitudinal}
\small
\begin{tabular}{rccccc}
\toprule
Step (M) & FC1 active & FC2-output PR & L1 PR & PCA95 & Reward $\pm$ SE \\
\midrule
1   & 44 & 4.34  & 6.29  & 16 & $-18.35 \pm 0.33$ \\
2   & 53 & 10.57 & 15.23 & 25 & $-11.37 \pm 0.52$ \\
5   & 62 & 16.98 & 23.43 & 40 & $+3.07  \pm 0.73$ \\
10  & 64 & 18.02 & 26.84 & 44 & $+9.82  \pm 0.43$ \\
20  & 64 & 22.35 & 34.27 & 46 & $+12.01 \pm 0.36$ \\
30  & 63 & 23.85 & 35.28 & 46 & $+13.34 \pm 0.33$ \\
50  & 64 & 26.11 & 39.13 & 48 & $+14.25 \pm 0.33$ \\
70  & 64 & 25.17 & 40.92 & 48 & $+13.08 \pm 0.33$ \\
90  & 64 & 29.72 & 40.32 & 49 & $+13.99 \pm 0.32$ \\
100 & 64 & 31.26 & 40.87 & 50 & $+14.76 \pm 0.28$ \\
\bottomrule
\end{tabular}
\end{table}

Three signatures commit on different timescales: \textit{which neurons participate} (FC1 active count, $\sim$10M), \textit{how well the policy plays} (reward, $\sim$20M), and \textit{how diverse the FC2-output representation is} (FC2-output PR, climbing through 100M without saturation). The high-PR basin is not committed early in this seed; it is climbed into. This refines the early-commitment story: active sets commit early in both b3 and t2, but PR-trajectory shapes differ across (game, regime, eval condition). $n = 1$ for the high-PR-climb pattern.

\section{V5/V6 Initialization Details}
\label{app:v5v6}

The only architectural difference between V5 and V6 Sample Factory experiments is the initialization of the policy and value heads:

\begin{table}[ht]
\centering
\caption{V5 vs.\ V6 initialization comparison. Only the policy and value heads differ.}
\label{tab:v5v6_details}
\small
\begin{tabular}{lll}
\toprule
Component & V5 Init & V6 Init \\
\midrule
CNN (frozen) & Kaiming uniform, seeded & Identical \\
FC layers & Orthogonal & Identical \\
Policy head ($64 \to 6$) & Kaiming uniform & Orthogonal, gain$=$0.01 \\
Value head ($64 \to 1$) & Kaiming uniform & Orthogonal, gain$=$1.0 \\
\midrule
Initial logit magnitude & 28--55 & 0.4--1.4 \\
Initial entropy & $\sim$0.0 (collapsed) & $\sim$1.7 (near-maximum) \\
\bottomrule
\end{tabular}
\end{table}

The V5 entropy collapse occurs because large initial logits produce near-one-hot softmax distributions, not because of the frozen CNN per se. SB3 uses orthogonal std$=$0.01 for the policy head by default, which is why SB3 frozen runs never exhibit entropy collapse.

\section{Physical Analogue: Lensless Imaging as a Random Projection}
\label{app:diffusercam}

The frozen-CNN bottleneck mechanism (Section~\ref{sec:mechanism}) has a direct physical analogue in computational imaging. In lensless cameras such as DiffuserCam \citep{antipa2018diffusercam}, a random optical diffuser replaces the traditional lens, scattering each scene point across the entire sensor in a physical random projection. The diffuser ``scrambles'' the incoming light, yet a computational reconstruction recovers a sharp image. The random scatter preserves geometry because intrinsically low-dimensional inputs (a finite scene) hit a high-dimensional sensor at random, and the high projection dimension $\gg$ the intrinsic scene dimension is exactly the Johnson--Lindenstrauss regime in which pairwise distances are preserved with high probability \citep{johnson1984extensions, dasgupta2003elementary}. Reconstruction is then possible because the sparse-in-some-basis scene admits a unique sparse solution under the compressed-sensing conditions of \citet{candes2006robust, donoho2006compressed}. The same principle has long been exploited in data science for random-projection-based dimensionality reduction \citep{bingham2001random, achlioptas2003database}.

The mapping to our setting is one-to-one. The frozen CNN plays the role of the optical diffuser: it produces a 3,136-dimensional random scattering of the pixel input. The 5-dimensional Pong manifold is the analogue of the finite scene. The trained FC readout plays the role of the computational reconstruction algorithm, recovering the sparse task-relevant directions of the projection. The key difference is that DiffuserCam's reconstruction is solved by an explicit convex program with $\ell_1$ regularization, while our setting recovers a sparse readout through gradient descent alone, without an explicit sparsity objective. Frozen-CNN reinforcement learning is, in this view, computational imaging of the task structure itself.

This analogy informed the framing of the mechanism but is not load-bearing for any quantitative claim in the main text. We include it here for readers from optics or signal processing who may recognize the random-scatter-plus-sparse-reconstruction architecture from their own field.

\section{Theoretical Context: Candidate Mechanisms}
\label{app:theory}

The standard theoretical result for gradient descent on $\min_W \|WX - Y\|^2$ is convergence to the minimum $\ell_2$-norm solution, which is generically dense \citep{gunasekar2018implicit}. Three candidate mechanisms may explain why we observe sparsity instead:

\textbf{Mechanism 1: ReLU-induced diagonal structure.} Each FC1 neuron computes $h_k = \text{ReLU}(\mathbf{w}_k^\top \phi(\mathbf{x}))$, creating effective diagonal structure. \citet{woodworth2020kernel} showed that diagonal linear networks with small initialization converge to minimum $\ell_1$-norm solutions. If ReLU creates analogous dynamics, this explains emergent sparsity.

\textbf{Mechanism 2: Large learning rate dynamics.} \citet{andriushchenko2023sgd} showed that large step sizes induce oscillation-driven sparsity through ``loss stabilization.'' Our learning rate of $2.5 \times 10^{-4}$ with linear decay may operate in this regime during early training.

\textbf{Mechanism 3: RL non-stationarity.} The RL loss landscape constantly shifts as the policy changes, potentially preventing settlement into the dense minimum-$\ell_2$-norm basin and deflecting toward sparser solutions.

These mechanisms are not mutually exclusive. Closing this gap rigorously remains important future work.

\section{Experimental Details}
\label{app:details}

\subsection{Game Details}

\begin{table}[ht]
\centering
\caption{Games studied and their estimated policy-relevant state complexity. Estimates are grounded in domain analysis and broadly consistent with AtariARI RAM annotations \citep{anand2019unsupervised}.}
\label{tab:games}
\small
\begin{tabular}{p{0.14\linewidth}p{0.42\linewidth}p{0.14\linewidth}p{0.08\linewidth}}
\toprule
Game & Policy-Relevant State Variables & Est.\ Vars & Actions \\
\midrule
Pong & Ball $(x,y,v_x,v_y)$, agent paddle $y$ & $\sim$5 & 6 \\
Breakout & Ball $(x,y,v_x,v_y)$, paddle $x$, brick pattern & $\sim$15--25 & 4 \\
Space Invaders & Ship $x$, alien grid, bullets, shields & $\sim$40+ & 6 \\
Freeway & Chicken $y$, 10 lane traffic phases/gaps & $\sim$10--15 & 3 \\
\bottomrule
\end{tabular}
\end{table}

\textbf{Pong} is a paddle game in which the agent returns a ball against an AI opponent; reward ranges from $-21$ (complete loss) to $+21$ (complete win). \textbf{Breakout} requires bouncing a ball off a movable paddle to clear a wall of bricks; each brick cleared earns one point (with reward clipping), and expert play scores 400+. \textbf{Space Invaders} requires shooting descending alien formations while dodging return fire. \textbf{Freeway} tasks the agent with guiding a chicken across ten lanes of traffic; car collisions push the chicken back but do not end the episode, enabling high-reward degenerate policies.

\subsection{Stochasticity Conditions}

\begin{table}[ht]
\centering
\caption{Stochasticity conditions across all experiments. ``Sticky'' indicates $p_\text{sticky}$, the probability that the previous action is repeated.}
\label{tab:sticky}
\small
\begin{tabular}{llcl}
\toprule
Game / Framework & Env ID & Sticky & Note \\
\midrule
Pong (SB3, 200M seeds) & \texttt{PongNoFrameskip-v4} & 0\% & Opponent reactive \\
Pong (SB3, 800M) & \texttt{PongNoFrameskip-v4} & 25\% & VecStickyAction \\
Pong (SF) & \texttt{ALE/Pong-v5} & 25\% & ALE-level \\
Breakout (SB3, Mac) & \texttt{BreakoutNoFrameskip-v4} & 0\% & Ball reactive \\
Breakout (SB3, Ubuntu) & \texttt{BreakoutNoFrameskip-v4} & 0\% & Ball reactive \\
Space Invaders (SB3) & \texttt{SpaceInvadersNoFrameskip-v4} & 25\% & VecStickyAction \\
Freeway (SB3, no sticky) & \texttt{ALE/Freeway-v5} & 0\% & Deterministic \\
Freeway (SB3, sticky) & \texttt{ALE/Freeway-v5} & 25\% & VecStickyAction \\
\bottomrule
\end{tabular}
\end{table}

\subsection{Frozen CNN Implementation}

The frozen CNN uses Kaiming normal initialization with a fixed seed. After initialization, all convolutional parameters have \texttt{requires\_grad} set to \texttt{False}. Bias terms are initialized to zero. FC layers use orthogonal initialization (SB3 and SF encoder layers).

\subsection{Evaluation Protocol}

All ablation evaluations use 50 episodes with deterministic (greedy) action selection. Activation statistics are computed from 1,000 frames sampled during evaluation.

\subsection{Computational Resources and Training Hyperparameters}

SB3 experiments: Apple Mac (M4 Max) at $\sim$1,700 steps/second (MPS), and Ubuntu desktop (AMD Ryzen 9 9950X, NVIDIA RTX 5090, 96GB DDR5 RAM) at $\sim$5,800 steps/second (CUDA, \texttt{SubprocVecEnv}). SF experiments: Ubuntu desktop at $\sim$21,000 steps/second (asynchronous APPO). Total compute: approximately 60 billion environment frames across all experiments.

SB3 hyperparameters: 8--128 parallel environments, rollout length 128, batch size 256, $n_{\text{epochs}} = 4$, clip range 0.1, entropy coefficient 0.01, $\gamma = 0.99$, GAE $\lambda = 0.95$, max gradient norm 0.5. Orthogonal initialization with std$=$0.01 for the policy head (SB3 default). SF hyperparameters matched where applicable, with Adam epsilon $10^{-6}$.

\subsection{Linear Probe Analysis}
\label{app:linear_probe}

To separate feature geometry from optimization dynamics, we fit $\ell_1$-regularized multinomial logistic regression predicting expert actions from frozen CNN features (Pong, 3,136 dimensions, $\sim$155k frames). With strong regularization ($C = 0.001$), the probe achieves 90.9\% accuracy using 1,091/3,136 features (65\% sparse). The $\sim$300$\times$ compression from linear probe ($\sim$1,000 features) to PPO ($\sim$3 neurons) reflects ReLU nonlinearity's ability to extract task-relevant subspaces more efficiently than linear methods.

\end{document}